\documentclass{article}
 \usepackage[preprint]{neurips_2025}

\usepackage[utf8]{inputenc} 
\usepackage[T1]{fontenc}    
\usepackage{hyperref}       
\usepackage{url}            
\usepackage{booktabs}       
\usepackage{amsfonts}       
\usepackage{nicefrac}       
\usepackage{microtype}      
\usepackage{xcolor}         
\usepackage{graphicx}

\usepackage{subcaption}
\usepackage{amsmath}
\usepackage{enumitem}
\usepackage{wrapfig}
\usepackage{algorithm}
\usepackage{algpseudocode}

\title{Avoiding Death through Fear Intrinsic Conditioning}

\author{Rodney Sanchez \\
  Dept. of Electrical and Computer Engineering\\
  Rochester Institute of Technology\\
  Rochester, NY 14626 \\
  \texttt{ras8047@rit.edu} \\
  \And
   Ferat Sahin \\
  Dept. of Electrical and Microelectronic Engineering \\
  Rochester Institute of Technology\\
  Rochester, NY 14626 \\
  \texttt{feseee@rit.edu} \\
  \AND
  Alexander Ororbia\\
  Dept. of Computer Science \\
  Rochester Institute of Technology\\
  Rochester, NY 14626 \\
  \texttt{agovcs@rit.edu} \\
  \And
  Jamison Heard \\
  Dept. of Electrical and Microelectronic Engineering\\
  Rochester Institute of Technology \\
  Rochester, NY 14626 \\
  \texttt{jrheee@rit.edu} \\
}

\begin{document}

\maketitle

\begin{abstract}
    Biological and psychological concepts have inspired reinforcement learning algorithms to create new complex behaviors that expand agents' capacity. These behaviors can be seen in the rise of techniques like goal decomposition, curriculum, and intrinsic rewards, which have paved the way for these complex behaviors. One limitation in evaluating these methods is the requirement for engineered extrinsic rewards for realistic environments. A central challenge in engineering the necessary reward function(s) comes from these environments containing states that carry high negative rewards, but provide no feedback to the agent. Death is one such stimuli that fails to provide direct feedback to the agent. In this work, we introduce an intrinsic reward function inspired by early amygdala development and produce this intrinsic reward through a novel memory-augmented neural network (MANN) architecture. We show how this intrinsic motivation serves to deter exploration of terminal states and results in avoidance behavior similar to fear conditioning observed in animals. Furthermore, we demonstrate how modifying a threshold where the fear response is active produces a range of behaviors that are described under the paradigm of general anxiety disorders (GADs). We demonstrate this behavior in the Miniworld Sidewalk environment, which provides a partially observable Markov decision process (POMDP) and a sparse reward with a non-descriptive terminal condition, i.e., death. In effect, this study results in a biologically-inspired neural architecture and framework for fear conditioning paradigms; we empirically demonstrate avoidance behavior in a constructed agent that is able to solve environments with non-descriptive terminal conditions.
\end{abstract}

\section{Introduction}
\label{sec:intro}
Real-world environments consist of many permanently damaging states that humans and animals learn to avoid without requiring them to experience these states explicitly. As a result, these beings are able to engage in safer exploration of their environments, such as by learning from the communicated experiences of others, reducing the need to sample the entire state-space. Moreover, many of these non-descriptive terminal conditions are related to concepts such as danger and death. Endowing autonomous systems, such as those in robotics, with the same ability to avoid precarious situations and agent-terminating conditions \cite{ororbia2023mortal} would prove immensely useful in constructing technological artifacts that better engage with human collaborators in a safe, viable manner.

Previous efforts in reinforcement learning (RL) that optimize against danger focus on control-based methods  \cite{malato2025search, 10645565, zhu2023importance} that require the agent to have an understanding of the environment's transition function in order to stop the agent from performing potentially dangerous or detrimental actions. 
Other work in RL \cite{zhou2025computationally} focuses on creating the environment transition function that slowly builds a representation of the dangerous actions that drives an agent to avoid these actions. A drawback of these approaches is that they require explicitly sampling dangerous states, which limits deployment on critical robotic systems since these states may damage the robot.

Pre-trained policies that are distilled to the learning agent have also been employed. Although these methods circumvent the need for explicit sampling, the fixed prior policy can only ensure that previously-encountered state representations are avoided \cite{zhu2023importance} but cannot further adapt to future situations. Furthermore, the uncertainty of avoidance that the previous policy learns makes it difficult to understand what stimuli the agent learned to avoid. Finally, these particular policy methods require a known optimal policy that can be used to initialize the agent, which is not always viable for robotic systems. 
Memory-based approaches center around using a single stimuli state, limiting the agent's ability to learn the relationship between the stimuli and the danger/threat \cite{sanchez2024fear}. This setup produces agents that seek to minimize the presence of the memorized stimuli, ultimately preventing the agent from exploring states close to the stimuli and resulting in undesired behavior; for instance, consider vehicular navigation, where other vehicles are a potential danger but we want the autonomous system to safely interact/coordinate with them and not completely avoid them. 
Prior approaches, such as the ones reviewed above, result in learning paradigms that require extensive prior knowledge or are dependent on insufficient stimuli representation(s) and often struggle to produce good policies, standing in stark contrast to the ease in which natural/biological entities would handle the same kinds of environments. 

In this study, we construct a low-shot learning RL system that engages safe, danger-avoidant behavior through a form of low-shot memory inspired by the type of social learning that humans use to learn avoidance. Specifically, this work introduces the siamese long-short term Memory (SLSTM) controller, which allows memory-augmented neural networks (MANNs) to learn from a series of states. This series of states mimics how humans undergo social conditioning by parents or peers since humans require detailed representations of the conditioning event(s). We introduce mechanisms, inspired by social conditioning, that uses state sequences to express a value that drives the agent to understand the relationship between the stimuli and itself. Our model demonstrates how mimicking social learning elements such as fear in young animals can allow an artificial agent to learn to avoid non-descriptive terminal conditions with limited representation information. 
In summary, the key contributions of this work are as follows:
\begin{itemize}[noitemsep,nolistsep]
    \item We propose a novel LSTM architecture that handles multimodal inputs at the cell/unit level; 
    \item We define and operationalize a behavioral scheme based on social conditioning for artificial control agents and craft a memory-augmented neural architecture that can handle image sequences inspired by this scheme; 
    \item We design a low-shot intrinsic reward functional that can deter the agent from non-descriptive terminal conditions without having to directly experience/sample them. 
\end{itemize}
Our proposed computational approach is studied and validated in the context of the Miniworld SideWalk problem, an extremely difficult partially observable Markov decision process (POMDP) with non-descriptive terminal conditions.
Our method takes an important step towards a single-life RL learning paradigm by providing a biologically inspired method that avoids terminal states. Something that single life RL will need when training in dangerous environments like the real world. 
\section{Related Work}
This work's focus on social fear warrants an interdisciplinary review that explores some pathological representations of fear. Our work focuses on the value of fear for RL agents, which may depict fear positively. However, we want to acknowledge that we understand that fear and its pathologies are detrimental to those afflicted by these disorders.
\noindent 
\textbf{Reinforcement Learning. } 
Reinforcement learning (RL) is an important sub-domain of statistical learning that focuses on optimizing the rewards and information related to an environment over time \cite{Sutton2018ReinforcementProgress}. 
In RL, different qualities of the environmental representations can change how the agent behaves to solve given tasks. Among those qualities is partial observability, which occurs when an agent's state no longer describes the full environment but only a portion of it \cite{DBLP:conf/nips/Doshi-Velez09}.
Other environmental conditions that can require specific adaptations include the delay in state space reward assignment, yielding often what is known as \emph{sparse rewards}. Here, the agent is only given a reward at the end of the episode, producing a long time horizon (e.g., the Atari games and the inverted pendulum problem \cite{brockman2016openaigym, Mnih2015}). This long horizon makes results in the temporal credit assignment problem, where it is difficult to assert which policy part led to the reward. 
Another reward paradigm in RL is known as the intrinsic reward; however, intrinsic rewards do not come from the environment and are instead produced by the agent's internal world model. These intrinsic rewards provide a reward signal to the agent when some internal condition is met, producing novel behaviors not dictated by the environment \cite{BartoIntrin} and encouraging a more intelligent form of exploration (beyond taking random actions). 

There have been two major intrinsic reward topics that have been studied in depth. The first topic promotes exploration of novel state representations \cite{Burda2019ExplorationDistillation,haarnoja2018soft,machado2020count,sekar2020planning,yao2021sample}. The second, in contrast, is focused on skill development and seeks to push the agent create and master sub-policies to solve the larger/overall goal,(i.e., the development of ``skills'' \cite{10109841, li2021learning}). Some research efforts, such as \cite{sanchez2024fear} and \cite{ODRIOZOLAOLALDE2025110055}, have designed negative forms of motivation for agents. In \cite{sanchez2024fear}, knowledge of memory-based states is utilized to provide approximate negative intrinsic reward values, while \cite{ODRIOZOLAOLALDE2025110055} uses prior knowledge of the environment's transition function. These reward variations fall under the field of \emph{reward shaping}, which seeks to create reward functions that promote successful completion of a particular task (often incorporating/encoding problem-specific domain knowledge into the shaped reward function), emulating or maintaining some restrictions. 

Some forms of reward shaping require on-policy learning methods (e.g., proximal policy optimization (PPO)) to solve a given task/environment \cite{schulman2017proximal}. On-policy learning is when an agent uses the knowledge of already visited future states to calculate the value of their current state. This process requires states to be ``within'' the policy to calculate the current state value, generally resulting in a less optimal policy than off-policy approaches \cite{Sutton2018ReinforcementProgress}. Offline methods (e.g. \cite{zhu2023importance}) utilize an off-policy formulation in POMDPs by guiding the policy to a known optimal policy. However, both on-policy and off-policy methods' value approximation is based on traditional conditioning paradigms that focus explicitly on learning about the environment through the direct visitation of states. To work effectively, these methods require a great deal of collected data to capture the entire representation of other agents' behaviors. As a result, these approaches struggle to integrate information from other agents' policies (limiting them to classical conditioning) and are ultimately unable to work with scenarios that include non-descriptive terminal conditions; such scenarios/problems remain underexplored and are only present in a few environments like ``Sidewalk'' and ``Lavagap'' \cite{Chevalier-Boisvert2024MinigridTasks}. 

\noindent 
\textbf{Memory Methods. } 
Memory methods in machine learning focus on encapsulating and recalling trained representations to allow a model to discern between inputs \cite{Hopfield1982-cp}.
Among the many models that have been developed over the years, memory-augmented neural networks (MANNs) have proven to be particularly useful for leveraging external memory for low/few-shot classification problems \cite{Santoro2016Meta-LearningNetworks}. This architecture used neural Turing machines to access memory \cite {graves2014neural}.
The MANN model was further expanded to work with convolutional network structures to acquire separable representations capable of distinguishing between different classes \cite{Mao2022}. Modern variants of MANN-like memory methods use transformers, e.g., taking advantage of their self-attention-driven key-value querying mechanism, as a means to further encapsulate representations. However, due to the great expense of training transformers, these memory-based approaches require large datasets and effectively struggle to perform low-shot learning \cite{10.1007/978-3-031-19815-1_37}. 
In this work, we will draw from and expand upon earlier, simpler formulations of MANN memory models to take advantage of their greater data efficiency and ability to incorporate prior representations to predict newly encountered stimuli. 

\noindent 
\textbf{Biological Fear. } 
Fear as a biological process is an automatic response that checks for the possibility of risk and danger, which falls within the brain's ``safety system'': composed of the amygdala, hippocampus, and frontal lobe \cite{MILAD2007446}. 
Notably, the amygdala computes the saliency of a given stimuli, which is then recalled from memory. This recall allows the agent to estimate its relative fear based on prior interactions. Critically, the frontal lobe also plays a role in fear inhibition, reducing the fear a stimulus exhibits after the amygdala produces its salience measure \cite{MILAD2007446}. Importantly, studies on general anxiety disorder (GAD) show that patients will encode irrelevant features of the stimuli. This over-encoding makes patients believe they are always at risk \cite{Laufer2016-qu}. 

For social animals, fear can be conditioned vicariously as agents share information about the stimulus \cite{DELGADO200639}. When this occurs, the agent will undergo conditioning automatically regardless of the agent's knowledge of the stimulus \cite{DESMEDT2015290,HERMANS2006361,MINEKA2002927}. When the fear is acquired, the agent can also undergo inhibition either by itself or by its peers, reducing the expected risk based on (already) known information \cite{JOVANOVIC20051559}. 

Vicarious conditioning \cite{Askew2008TheOn} itself can be represented in four steps: 
1) attention, which describes what the agent learns from its peer; 
2) retention, which describes how much information is stored; 
3) major reproduction, when the agent applies the behavior to its environment; and, finally; 
4) reinforcement, where the agent confirms the usefulness of the behavior. 

For a collective of social animals, this type of fear social conditioning includes modifying the fear by their social group, which means that animals can condition each other by relaying information about the fear stimulus \cite{Marin2020VicariousDyads}. Critically, the parents serve as accurate conditioning peers for the younger animals\cite{Sullivan2020-ho}. Here, the young offspring will trust the parent until adolescence, after which social learning typically switches to their peer(s) for conditioning (the caveat is that the peer's information is modulated by the animal's trust in the peer) \cite{Sullivan2020-ho}. Note that conditioning occurs even if the agent knows that the stimulus is not dangerous; this happens because neither fear acquisition nor extinction can be consciously halted or (re)started \cite{exposure_therapy_fear}. This differs from inhibition, which describes an active process of fear rationalization for modifying the stimulus's predicted risk \cite{fear_memory_psycology_treatment}. This explicit priority of subconscious prediction of risk, then conscious modulation, puts priority on risk avoidance. This process allows inhibition to explore the stimuli, which, if miscalculated, can lead to fear extinction (i.e., unlearning a learned fear response) \cite{JOVANOVIC20051559}.

Social fears, automatic conditioning, and peer inhibition collectively allow agents to maximize their stimuli knowledge, averaging the negative value of peer conditioning with the positive peer inhibition so as to create a more accurate representation of the stimulus's value prior to sampling. This is imperative with respect to stimuli that may lead to death or permanent debilitating injuries. 

\section{Methodology}
\label{sec:methodology}

Fear conditioning paradigms in humans depend on social learning, where discussing fear stimuli can serve as a conditioning event for the observer or listener. Here, the agent learns of negative behavior by memorizing the example(s) shared with them. The agent then compares its behavior to the one it retained (in memory) to produce an approximation of the value of its behavior. The ability to store a series of steps as opposed to a singular representation is at the core of social conditioning, as it describes how a behavior can cause an environment to produce a negative reward signal. Behavior, unlike policy, differs in that it describes a series of state transitions that share a total negative reward. 

To formally describe social conditioning, we first provide a concept of behavior in Equation \ref{behavior_parent}. We borrow the term ``behavior'' from Skinner's behaviorism, which describes the alteration of internal states in response to stimuli. Specifically, we denote behavior as a set of state transitions, regardless of the agent's actions, emulating Skinner's expectation of internal states (i.e., fear, responding to stimuli), which results in changes in the environment's states. We then define the social behavior value with the key difference being that, unlike a policy that defines an explicit set of actions or transitions with a known value of each, behavior gives the agent a \emph{collective value} for a set of state transitions, irrespective of the actions that the agent took. A behavior $\beta$ will have a fixed value for all transitions given some other agent that provided the behavior. Furthermore, we define the value of a behavior differently from the regular value function due to its explicit time horizon and predefined path: 
\begin{align}
\{ &P^{\pi}(a_{1}|S)P(S'|a_{1},s),...P^{\pi}(a_{n}|S)P(S'|a_{n},s), \notag \\
&P^{\pi}(a_{1}|S')P(S''|a_{1},s'),...P^{\pi}(a_{n}|S)P(S''|a_{n},s')\} \subset \beta_{parent} \label{behavior_parent}
\end{align} 
\begin{equation}
 V\beta_{Parent}=V{\pi}_{Parent}(S'')P(S''|*,S)+V\pi_{Parent}(S')P(S'|*,S)+V\pi_{Parent}(S)
 \label{value_parent}
 \end{equation} 
 where $a_i$ is the $i$-th action taken by the parent, $s \in S$ is the current state, $s^\prime \in S^\prime$ the next state, and $s^{\prime\prime} \in S^{\prime\prime}$ is the state after the next state.

\begin{figure}[!t]
    \centering
    \includegraphics[width=0.8\textwidth, trim={0 0 0 0.7cm}, clip]{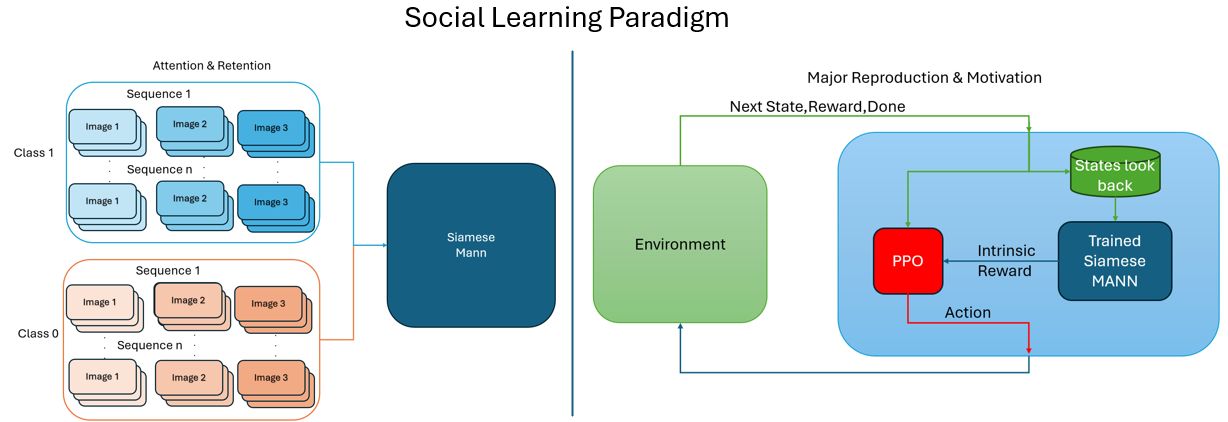} 
    \caption{Depicts the social learning frameworks where attention and retention use low-shot learning to train the Siamese MANN while reproduction and motivation occur through intrinsic rewards.}
    \label{Full_Social}
\end{figure}

Important to social learning paradigms is the degree of ``trust'' or the assured value of behavior; this is based on how much the learning agent believes the peer knows about the environment. A more general version of the value of the behavior, in the context of a peer(s), is as follows: 
\begin{align}    
\{P^{\pi}(a_{1}|S)P(S'|a_{1},s),P^{\pi}(a_{2}|S)P(S'|a_{1},s)...P^{\pi}(a_{n}|S)P(S'|a_{n},s), \notag \\ 
P^{\pi}(a_{1}|S')P(S''|a_{1},s'),P^{\pi}(a_{2}|S')P(S''|a_{1},s')...P^{\pi}(a_{n}|S)P(S''|a_{n},s') \} \subset \beta_{peer}  \label{behavior_peer}
\end{align}
\begin{equation}
      V\beta_{Peer}=V{\pi}_{Peer}(S'')P(S''|*,S)\kappa_{S''}+V\pi_{Peer}(S')P(S'|*,S)\kappa_{S'}+V\pi_{Peer}(S)\kappa_{S}
      \label{value_peer}
\end{equation}
where $\kappa_{S^{*}}$ describes a reliability factor of the peer at a given state. This factor is difficult to isolate as it can be culturally defined; the reliability given peer information can be derived as a prior utility or something more abstract, e.g., ``agreeableness''. Since our work is inspired by early amygdala learning, we treat the peer-presented data as a parental figure and therefore assume that $\kappa_{S^*} = 1$.  

The entire conditioning method can be seen in Figure \ref{Full_Social}. Important to social learning paradigms is the consideration that the value is not fixed to a predefined state but instead to the state representation of the peer/parent. Since we cannot know the peer/parents' state representation, we assume that the agent has to recall memory representations that best describe that state. To create a memory method that understands these state transitions, we need a model that can extract feature-rich representations of the state and an input (image) while maintaining the information of previous reads and the state used by the MANN. To manage this mixed representation, we introduce the Siamese LSTM module.

\noindent 
\textbf{Siamese Long Short-Term Memory Module.} Current memory methods use an LSTM to learn an encoding / representation of state as well as handle previous reads and computed prior states. To accommodate the need to process stimuli (images) and prior reads (neural activity vectors), we 

\begin{wrapfigure}{l}{0.5\textwidth}
  \centering
  \includegraphics[width=0.49\textwidth, trim={0 0 0 0.5cm}, clip]{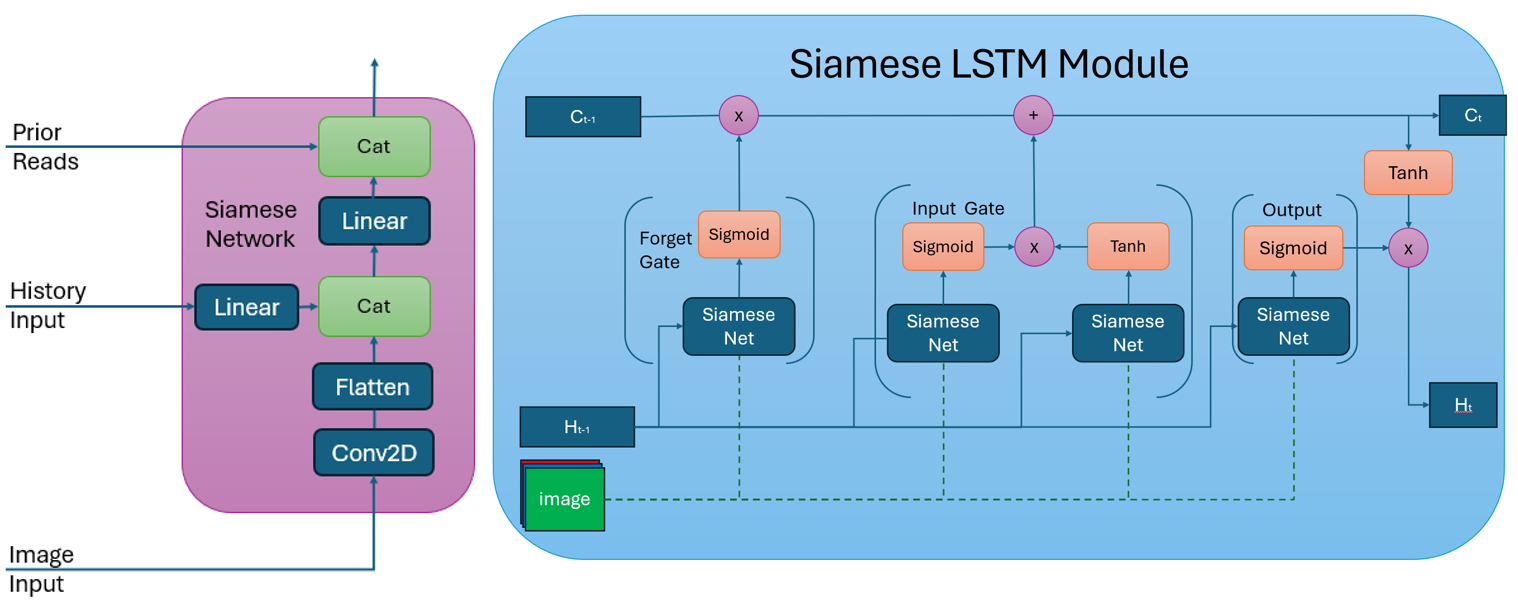}
  \caption{Demonstrates the Siamese network that constitutes the gate in our SLSTM Module, the inclusion of this network for the gates allows for the mixing of images and vectors. For a larger version, see the appendix.}
  \label{SLSTM_CELL}
\end{wrapfigure}
\begin{figure}[!t]
\begin{subfigure}[t]{0.48\textwidth}
  \centering
  \includegraphics[width=0.9\textwidth, trim={0 0 0 0.75cm}, clip]{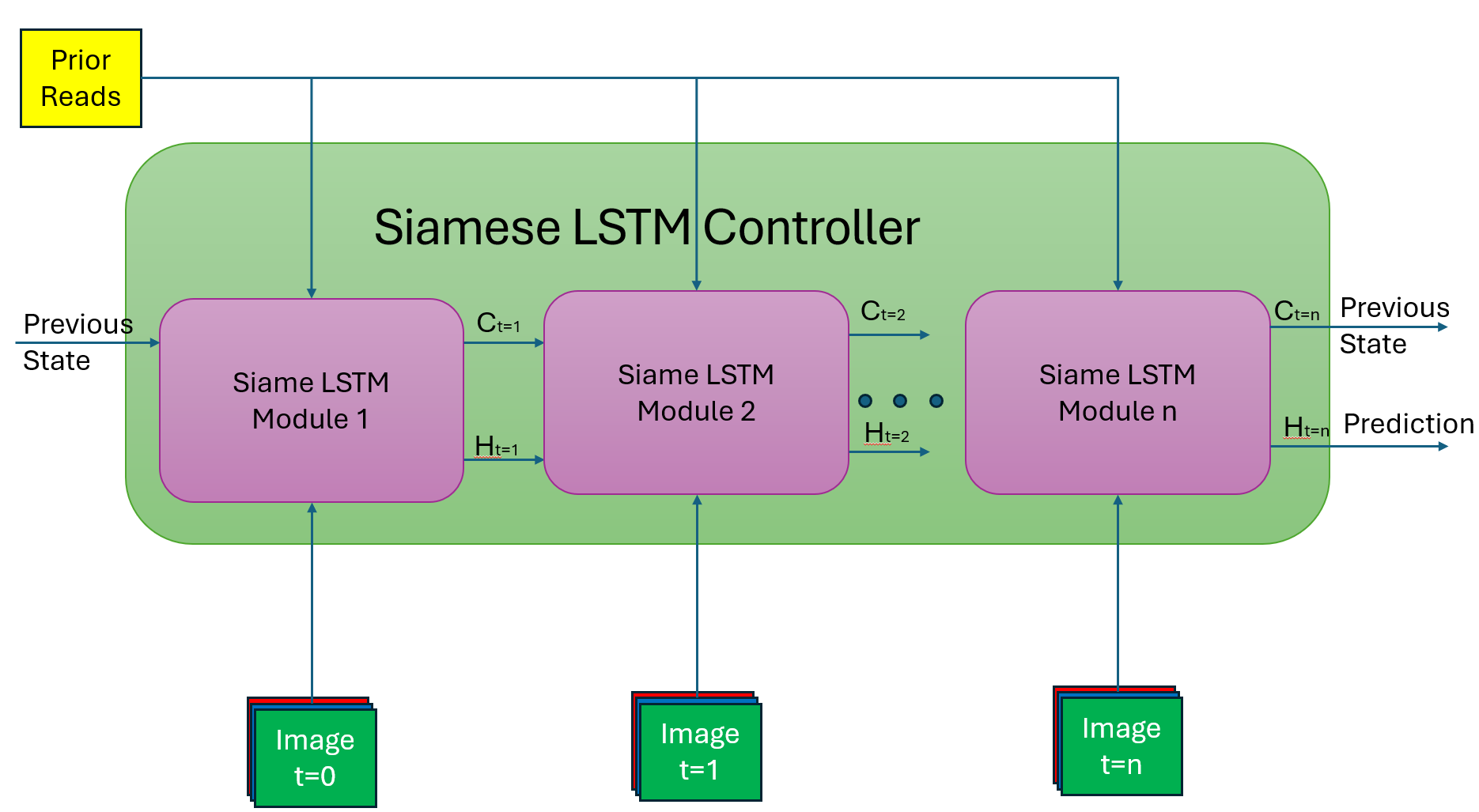}
  \caption{The SLSTM, which takes in an image sequence, prior state, and prior reads, to produce an encoding.}
  \label{SLSTM_COntroller}
\end{subfigure}
~
\begin{subfigure}[t]{0.48\textwidth}
  \centering
  \includegraphics[width=0.9\textwidth, trim={0 0 0 0.25cm}, clip]{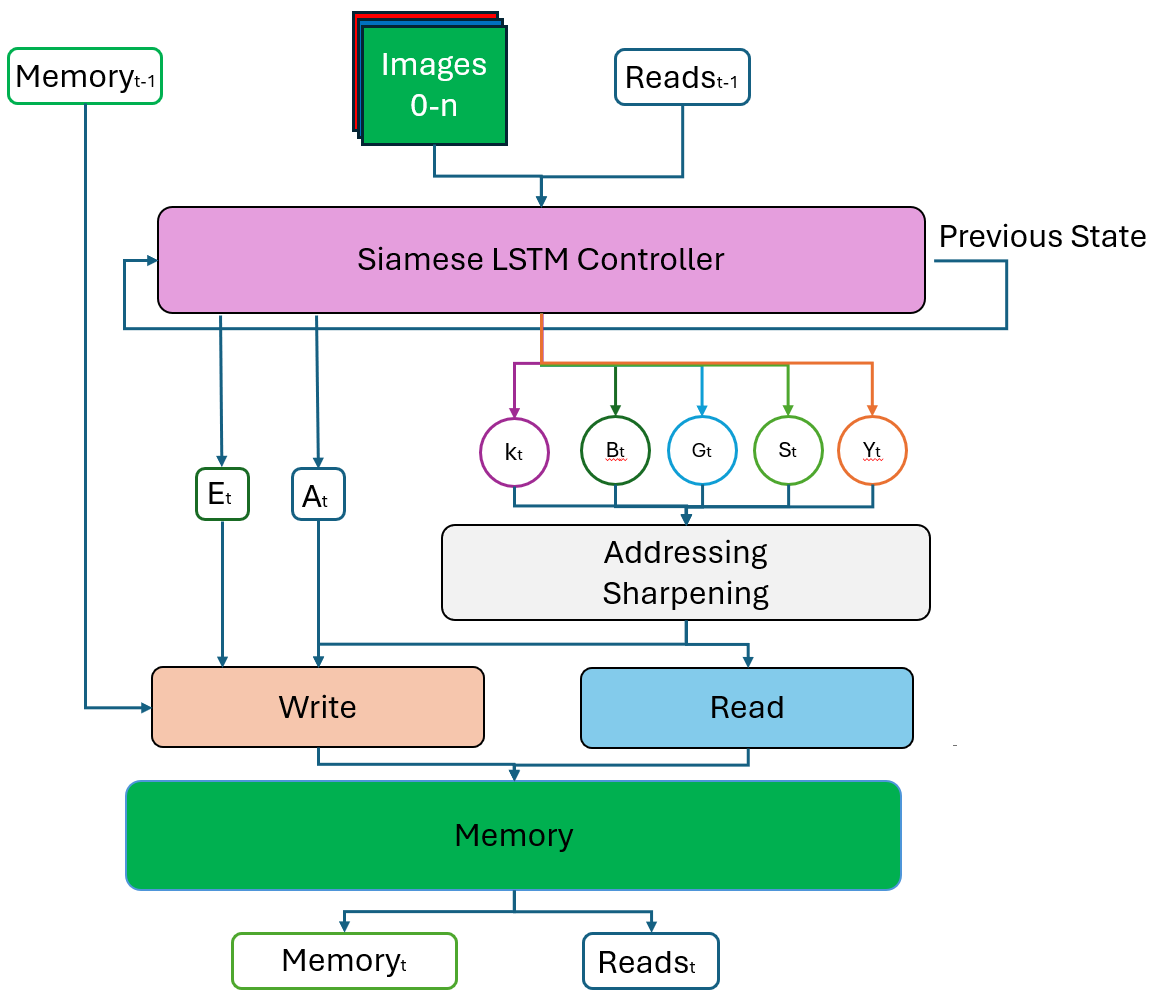} 
  \caption{A depiction of our full module, including the Siamese memory-augmented network component. }
  \label{Siamese_MANN}
\end{subfigure}

\caption{An overview of the SLSTM controller and MANN (see appendix for a larger version).}
\vspace{-0.6cm}
\end{figure}


propose the Siamese LSTM (SLSTM) module (see Figure \ref{SLSTM_CELL}). The Siamese LSTM module facilitates modality mixing within its internal states using a Siamese (or twin) neural network to replace the usual linear projection layer. This allows the recurrent network to act as a controller that can extract rich features of images while maintaining prior reads in order to produce the encoding. This inclusion creates a modular paradigm where different modality types, e.g., text, audio, or images, can be included into the controller to create a succinct memory encoding. Furthermore, the modularity of the Siamese net facilitates the extraction of features for each modality; the resulting many-to-one LSTM controller architecture allows the controller to compose an encoding from a sequence of images and vectors. This many-to-one controller is depicted in Figure \ref{SLSTM_COntroller}. We remark that, although we empirically observed no issues with our model(s) in this work, training might become difficult for some problem instances due to vanishing gradients (which can hinder model parameter optimization) given that the LSTMS, though highly-effective, can still struggle to capture longer-term dependencies over long temporal sequences (as a result of using backpropagation through time). 

\noindent 
\textbf{Siamese Memory-Augmented Neural Networks. } 
Memory is a core part of social conditioning since it plays a role in the recall process. It helps the agent reproduce the peers'/parents' intended behavior 
given the current example it is being conditioned on. Specifically, reading from its memory, the agent is able to produce a relatively known value based on how similar its current behavior is to that of its peers/parents. We model the underlying similarity calculation using cosine similarity when the MANN reads from memory; the agent receives an intrinsic reward proportional to the likelihood of the key (vector) being part of the peer(s)/parent(s) representation and the value (vector) of the peer(s)/parent(s). The integration of our SLSTM with the MANN results in the Siamese MANN (SMANN) for comparing behavior(s) (see Figure \ref{Siamese_MANN}). Since behavior is defined as a set of state transitions, the SLSTM's component drives the MANN to process the previous reads and the image sequence; it specifically processes the previous state, a set of state transitions, and previous reads, yielding an encoding that best describes the controller. This means that the agent can be provided with a small set of representations that encapsulates the parent's behavior and a known value for the behavior. The full SMANN is then trained on the $\beta_{parent}$, operationalizing the attention and retention dimensions of social learning. For major reproduction and reinforcement, the agent samples its environment and calculates the similarity between current states and its memory; a softmax predicts this similarity and dictates the probability of the behavior while also serving as an intrinsic reward. In the appendix, we provide an algorithmic depiction for the process of our SMANN.

Formally, we depict how our similarity is computed, in contrast to the traditional MANN, below: \\
\noindent\begin{minipage}{.4\linewidth}
\vspace{-0.25cm}
\begin{equation}
  K_{t}=\frac{LSTM(S_{t})\bullet M_{t}(i)}{||LSTM(S_{t})|| ||M_{t}(i)||}  
  \label{base_recall}
\end{equation}
\end{minipage}
\noindent\begin{minipage}{.6\linewidth}
\begin{equation}
  K_{t}=\frac{SLSTM(S_{t},S_{t-1},S_{t-2})\bullet M_{t}(i)}{||SLSTM(S_{t},S_{t-1},S_{t-2})|| ||M_{t}(i)||}  
    \label{our_recall}
\end{equation}
\end{minipage}
where $\bullet$ denotes the dot product and $||\cdot||$ denotes the Euclidean norm. As can be seen above, the integration of the SMANN changes the read operation from comparing similar states (Equation \ref{base_recall}) to comparing behaviors (Equation \ref{our_recall}). The negative intrinsic reward punishes any actions whose transitions simply mirror the behavior, pushing the agent to uncover a policy that does not collapse to mimicking the peers'/parents' policy. 

Finally, we explore modifying a threshold before the intrinsic reward is applied. Since the agent and parent do not share the same feature extractor, modifying the threshold modifies the allowable polices that the agent can create based on its behavior. A low threshold punishes any behavior that resembles the $\beta_{parent}$ while a high threshold only punishes behaviors that closely represent the dataset. Variation in encoding has been demonstrated in humans, where ``over-encoding'' is theorized to cause GAD. An abstraction that we believe this mechanism embodies is when a parent warns their child, ``Don't touch the hot Stove'' -- this pushes the child to mimic the behavior that the parent seeks to condition, not based on the policy implied by the literal actions. 
 
This form of learning is ideal for non-descriptive negative states, which lack an incremental representation of danger, e.g., threats such as poisonous plants and fast highways would not modify the agent's representation to elicit danger. Our thresholding scheme allows the agent to be deployed in a more realistic environment while minimizing the exploration of dangerous states. 

\section{Experimental Setup}
\label{sec:experimental_setup}

All experiments were conducted using MiniGrid's Sidewalk environment in order to evaluate our intrinsic fear paradigm. The environment employed used the default number of steps, yielding episodes with a maximum length of $150$ steps. Notably, the Sidewalk environment punishes the agent for all steps, reducing the maximum extrinsic reward by the step count over the maximum steps. All agent runs were trained for $1000$ episodes using a proximal policy, updating the agent every $450$ steps. Collectively, all of the experiments were run on an NVIDIA $3090$; the agent hyperparameter configuration used can be found in the appendix. The fear behavior dataset was produced by starting the agent at a non-fear state, then, within three actions, the agent was made to approach the terminal condition. For the non-danger class, the agent was randomly spawned and took three actions in any direction except towards the terminal condition. All stimuli datasets were created using the last state from its equivalent behavior dataset. Crucially for the stimuli, this means that the last state for the fear class faces towards the danger while the non-danger class faces away; this mimics how the stimuli dataset was previously created in order to train fear-based stimuli rewards. Both sets contain $38$ representations of each class.
Since we cannot ascertain the specific features that encompass the $\beta_{parent}$, we test what a variation of the features is by adjusting the threshold that the classifier needs to produce so that the negative intrinsic reward takes effect. Doing this allows us to test how ``stringent-ness'' of the danger/non-danger class representation affects our agents. Specifically, the threshold (for behavioral fear) was varied from $0.25 $to $0.95$ in order to observe how the increase in threshold modified the agent's acquired policy. Low thresholds ranged from $0.25$ to $0.45$, mid(dle) thresholds ranged from $0.5$ to $0.75$, and high thresholds ranged from $0.75$ to $0.95$. 
The memory-augmented neural network and the Siamese MANN were trained on the set with a batch size of two for $300$ epochs (MANN and Siamese MANN meta-parameters can found in the appendix). The developed code will be provided here: https://anonymous.4open.science/r/Behavior-Intrinsic-Fear-B76C. 

\section{Results}
\label{sec:results}
This section will discuss the results of altering the $\beta$ fear conditioning value, baseline PPO, and stimuli-based fear conditioning. Prior offline adaptation methods have achieved $0.769 \pm 0.060$ results, but these used predefined optimal policies \cite{malato2025search}. Since our method does not focus on offline polices, we do not compare directly against them but view them as a theoretical upper bound on how well RL methods could perform. In these results, we will look for successful solutions to the environment and increasing avoidance of terminal states. The overall trend observed across the results (Table \ref{general_trend_table}) indicates that a higher threshold helps the agent to better solve the environment with a greater probability of sampling a terminal state. Note that PPO alone cannot solve the environment due to its inability to receive information about the terminal state, and our method discourages the agent from exploring states that approximate the $\beta_{parent}$. This avoidance incentivizes the agent to uncover paths that avoid these states, which produces longer paths while decreasing the overall reward.

\begin{table}[!t]
  
  \centering
  \begin{tabular}{lllll}
    \toprule             
    \cmidrule(r){1-2}
    Method &Episode length &Intr Rew & Ext Rew & Max Ext Rew\\
    \midrule
    Base PPO &115.061$\pm{}$7.675 &0 &0  &hold \\
    PPO $\&$ Stimuli\cite{sanchez2024fear} &60.810$\pm{}$18.502 &-29.588$\pm{}$9.134  &0.000$\pm{}$0.001 &0.028 \\
\\
    PPO $\&$ low $\beta$=.25 &128.701$\pm{}$9.124 &-6.194$\pm{}$1.280  &0.001$\pm{}$0.002 &0.057 \\
    PPO $\&$ mid $\beta$=.60 &135.42$\pm{}$10.898 &-3.619$\pm{}$3.780  &0.000$\pm{}$0.001 &0.026    \\
    PPO $\&$ high $\beta$ =.95&107.952$\pm{}$6.524 &-9.546$\pm{}$4.504  &0.001$\pm{}$0.003 &0.082
\\ \midrule
    Agent $\&$ Offline &NA &NA  &0.769$\pm{}$0.060 &0.820
\\
    \bottomrule
    \end{tabular}
  \caption{Average episodic intrinsic and extrinsic reward across all methods. Results averaged across $5$ runs and $1000$ episodes. \textbf{Note:} ``Ext Rew'': extrinsic reward, ``Intr Rew'': intrinsic reward.} 
  \label{general_trend_table} \vspace{-0.5cm}
\end{table}

\begin{figure}[!h] 
   \centering
    \begin{subfigure}[t]{0.32\textwidth}
        \centering
        \includegraphics[width=0.9\textwidth, trim={0 0 0 1.2cm}, clip]{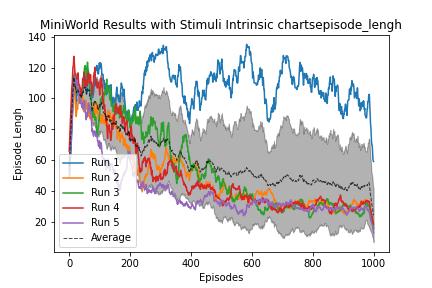}
        \caption{Episode Length}
        \label{fig:STIMULI_LENGHT}
    \end{subfigure}
~
    \begin{subfigure}[t]{0.32\textwidth}
        \centering
        \includegraphics[width=0.9\textwidth, trim={0 0 0 1.1cm}, clip]{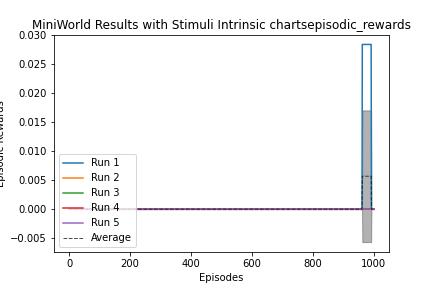}
        \caption{Episodic Reward}
        \label{fig:STIMULI_EXTRINSIC}
    \end{subfigure}
    ~
    \begin{subfigure}[t]{0.32\textwidth}
        \centering
        \includegraphics[width=0.9\textwidth, trim={0 0 0 1.2cm}, clip]{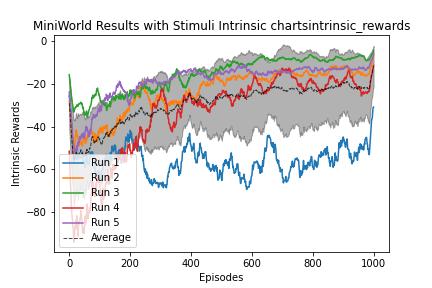}
        \caption{Episodic Intrinsic}
        \label{fig:STIMULI_INTRINSIC}
    \end{subfigure}

    \caption{The stimuli fear's constant negative reward approximates a living cost, making it optimal for the agent to go to the terminal condition, promoting a decrease in episode length. For a larger version, see the appendix.}
    \label{fig:STIMULI_RESULTS}
\end{figure}

\begin{figure}[htbp]
    \centering
    \medskip
    \begin{subfigure}[t]{0.32\textwidth}
        \centering
        \includegraphics[width=0.9\textwidth, trim={0 0 0 1.2cm}, clip]{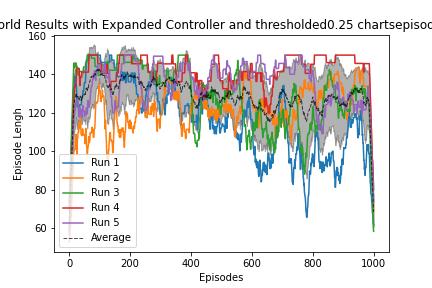}
        \caption{Episode Length}
        \label{fig:LOW_LENGHT}
    \end{subfigure}
    ~
    \begin{subfigure}[t]{0.32\textwidth}
        \centering
        \includegraphics[width=0.9\textwidth, trim={0 0 0 1.2cm}, clip]{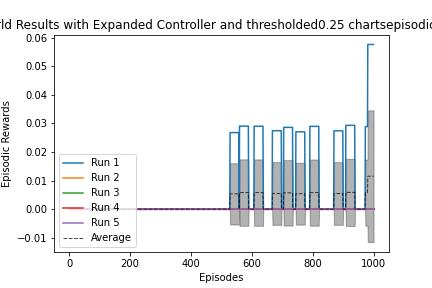}
        \caption{Episodic Reward}
        \label{fig:LOW_EXTRINSIC}
    \end{subfigure}
    ~
    \begin{subfigure}[t]{0.32\textwidth}
        \centering
        \includegraphics[width=0.9\textwidth, trim={0 0 0 1.2cm}, clip]{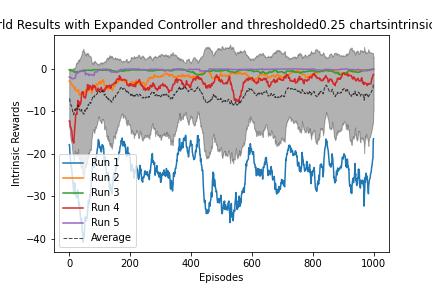}
        \caption{Episodic Intrinsic}
        \label{fig:LOW_INTRINSIC}
    \end{subfigure}

    \caption{At a low threshold value, the intrinsic reward punishes any policy that approximates the behavior promoting the longest episode length. For a larger version, see the appendix.}
    \label{fig:LOW_RESULTS}
\end{figure}

\begin{figure}[htbp]
    \centering
    \medskip
    \begin{subfigure}[b]{0.32\textwidth}
        \centering
        \includegraphics[width=0.9\textwidth, trim={0 0 0 1.2cm}, clip]{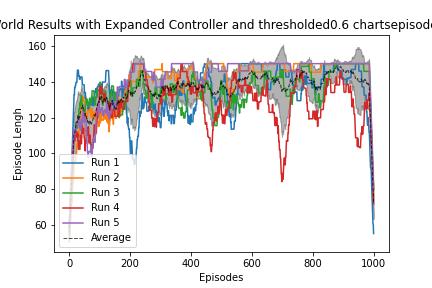}
        \caption{Episode Length}
        \label{fig:MID_LENGHT}
    \end{subfigure}
    ~
    \begin{subfigure}[b]{0.32\textwidth}
        \centering
        \includegraphics[width=0.9\textwidth, trim={0 0 0 1.2cm}, clip]{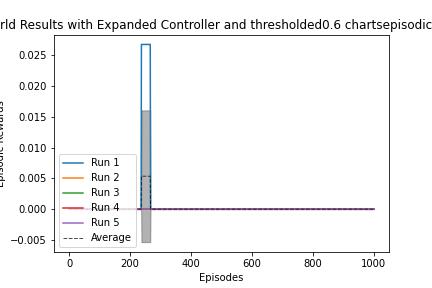}
        \caption{Episodic Reward}
        \label{fig:MID_EXTRINSIC}
    \end{subfigure}
    ~
    \begin{subfigure}[b]{0.32\textwidth}
        \centering
        \includegraphics[width=0.9\textwidth, trim={0 0 0 1.2cm}, clip]{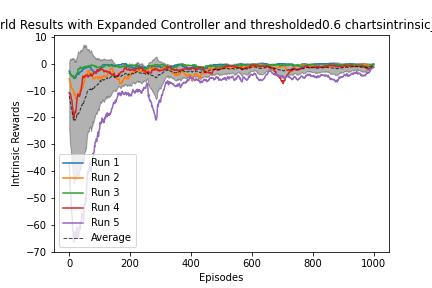}
        \caption{Episodic Intrinsic}
        \label{fig:MID_INTRINSIC}
    \end{subfigure}

    \caption{A mid(dle) threshold fear value promotes the learning of a policy that neither optimizes for greater episode length via safe exploration nor greater exploration via riskier actions. For a larger version, see the appendix.}
    \label{fig:BEHAVIOR_MID_RESULTS}
\end{figure}

The stimulus-based episode lengths and (trial-averaged) reward curves are depicted in Figure \ref{fig:STIMULI_RESULTS}, where the method will always produce a negative intrinsic reward (albeit small). This constant negative reward approximates a living cost, making it optimal for the agent to terminate the game even if it does find the goal.  
Furthermore, the results for low fear thresholds are provided in Figure \ref{fig:LOW_RESULTS}. Since a lower threshold will punish any state transitions that closely resemble $\beta_{parent}$, we expect it to be the least likely to find the goal yet produce the largest episode length. Our results demonstrate that the low threshold does produce the largest number of steps in general. Conversely, when it finds the goal, the negative reward incentivizes the agent to reach the goal. We also viewed a lower threshold as loosely mimicking the over-encoding of stimuli expected in GAD; our results verify that the over-encoding does promote extreme risk aversion. We visualize this behavior in the appendix. 
Our results for the mid(dle) threshold value are shown in Figure \ref{fig:BEHAVIOR_MID_RESULTS}. From the results, we see that middle threshold produced a larger episode length as compared to the low threshold; this occurred because the low threshold model was consistently able to find the goal whereas the middle threshold only found it once for a few episodes. This leads us to believe that a mid-threshold model would likely achieve a smaller episode length from the increased sampling of riskier states. 

\begin{figure}[htbp]
    \medskip
    \begin{subfigure}[b]{0.32\textwidth}
        \centering
        \includegraphics[width=0.9\textwidth, trim={0 0 0 1.2cm}, clip]{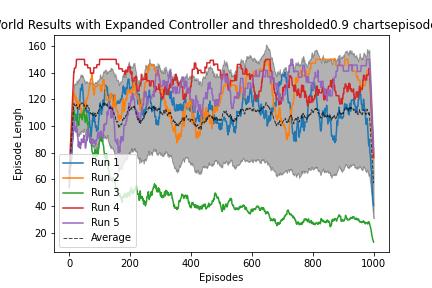}
        \caption{Episode Length}
        \label{fig:HIGH_LENGHT}
    \end{subfigure}
    ~
    \begin{subfigure}[b]{0.32\textwidth}
        \centering
        \includegraphics[width=0.9\textwidth, trim={0 0 0 1.2cm}, clip]{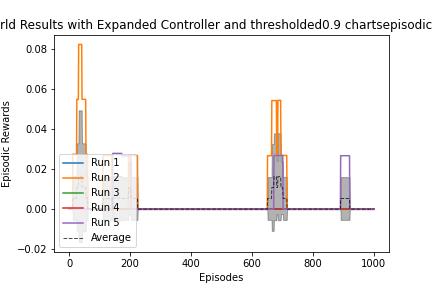}
        \caption{Episodic Reward}
        \label{fig:HIGH_EXTRINSIC}
    \end{subfigure}
   ~
    \begin{subfigure}[b]{0.32\textwidth}
        \centering
        \includegraphics[width=0.9\textwidth, trim={0 0 0 1.2cm}, clip]{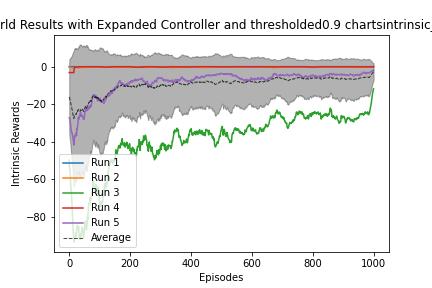}
        \caption{Episodic Intrinsic}
        \label{fig:HIGH_INTRINSIC}
    \end{subfigure}

    \caption{The high fear threshold model engages in the most exploration, increasing returns while still punishing the pristine representation of fear, further increasing the probability of success yet reducing episode length. For a larger version, see the appendix.}
    \label{fig:BEHAVIOR_HIGH_RESULTS}

\end{figure}

Finally, the results for a high threshold model are shown in Figure \ref{fig:BEHAVIOR_HIGH_RESULTS}. We expect that the increase in the threshold will also increase how often the agent finds the goal, yet decrease episode length from riskier behavior. This is supported by our results, which show that the agent finds the goal a greater number of times than prior thresholds while achieving a lower episode length. 

\noindent
\textbf{Analysis. } Overall, the results show that our $\beta$ fear model does a better job at mimicking socially learned fear since it discourages a specific unsafe behavior. A lower threshold (resilience to fear) did keep the agent safer; however, it limited the agent's capacity significantly. This outcome correlates with prior psychological theories that posit that GAD is an over-generalization of fear features. 
Our work permits controlling those variables and was found to elicit similar results, showing a correlation between the psychological theory and our method. Thus, our biologically-inspired method may better allow agents to exhibit some human tendencies.

We note that the proposed $\beta$-fear conditioning method extended the agent's life and promoted avoidance behaviors, resulting in a paradigm that introduces risk representations to the agent without requiring direct sampling or specific pre-trained policies. This approach has applications ranging from single-life RL to robotics or other domains where terminal conditions may have irreparable consequences. However, there exists an interesting relationship between the threshold (``resilience'') at which an agent experiences fear and the episode length, where too high of a fear threshold incurs more risk-taking behavior, i.e., not finding the goal as often. This result is attributed to the agent not generalizing the fear representations well, which means that the agent requires an extremely specific fear representation. Conversely, too low of a threshold ``traumatizes'' the agent, where the agent never approaches the fearful states, resulting in longer episode lengths and finding the goal state less often. These behaviors are similar to human behaviors where humans/animals may ``freeze up'' in fearful scenarios or fail to recognize imminent danger.  

There were instances where some thresholds failed to find the goal -- we attribute this to PPO lacking a proper in-built exploration paradigm. PPO without fear never finds a goal state and always fails to solve the environment. Thus, including fear helps limit the exploration area, stopping the agent from transitioning into non-descriptive terminal conditions. In cases where the agent does experience an extrinsic reward, the on-policy algorithm does appear to optimize for the extrinsic rewards and find a goal state. However, there is a large variance in the measured intrinsic rewards across runs for each threshold; we attribute this to the fact that the agent (re)spawns at random positions for each seed. Greater negative intrinsic rewards are produced when an agent spawns closer to the sidewalk or faces the sidewalk, which impacts the overall learned policy. 

\noindent 
\textbf{Ethical Analysis.} Behavioral fear does not explicitly dictate what features are used to learn a representation of a fear stimulus. This makes it essential for researchers to take care when using the proposed method in tasks that involve humans. An unbalanced dataset or including unaccounted-for bias towards any distinct group of people may further entrench that bias into the agent's learned behavior. Although this form of bias conditioning can occur among humans, humans do have greater, though not fully understood, capacities to overcome these biases. Furthermore, fear conditioning methods should not be deployed in critical infrastructure environments since an incorrectly-set threshold may lead to unexpected or undesired risk-taking behavior in the agent.

\noindent
\textbf{Limitations and Future Work. } 
One limitation of this work arises from the fact that Sidewalk's POMDP and sparse reward function promote using on-policy methods like PPO. However, PPO's lack of exploration makes it less likely to find the goal (despite our fear-conditioning helping in the task studied). Another limitation is the possible misalignment of the agents' feature extractor and the SMANN feature extractor -- this misalignment can lead the agent to avoid, inhibit, and recall different states based on its feature extractor. Humans represent the world through data received from sensory organs and maintain a shared feature space as a result of the hippocampus, amygdala, and prefrontal cortex.  In the biological community, different aspects of the safety system/network undergo what the machine learning community would call ``freezing'', where different parts of the safety network/system are updated/trained. This freezing and retraining paradigm can be considered for future work to maintain a common feature space. Finally, our method only mimics parent social conditioning with a trust ($\kappa$) value of one; this was a simplification made in this work, but future efforts should explore non-parental social conditioning with adaptive trust values.
 
\vspace{-0.25cm}
\section{Conclusion}
\label{sec:conclusion} 

This work demonstrates how a psychologically-inspired social fear learning paradigm can discourage an agent from exploring non-descriptive terminal states using low-shot learning mechanisms. The introduced memory module works to encode behaviors that deter the overall agent from a set of negative behaviors. Furthermore, this form of recall builds a more complete representation of social fear learning, as our results for a POMDP environment with non-descriptive terminal conditions indicate. A lower fear threshold correlated with generalized anxiety disorder, showing a promising overlap between our method and some psychological disorders. Overall, our $\beta$ fear conditioning framework demonstrates that a fear-based intrinsic reward can deter undesired behaviors in a low-shot manner. This method could prove to be a first step to realizing single-life RL systems in realistic environments where death, i.e., non-descriptive terminal state(s), is present.

\bibliographystyle{plain}
\bibliography{behavioral_fear_refrences}

\newpage

\section{Appendix / supplemental material}

\subsection{Larger Methodology Figures}
\begin{figure}[H]
\begin{subfigure}[t]{\textwidth}
  \centering
  \includegraphics[width=\textwidth, trim={0 0 0 0.75cm}, clip]{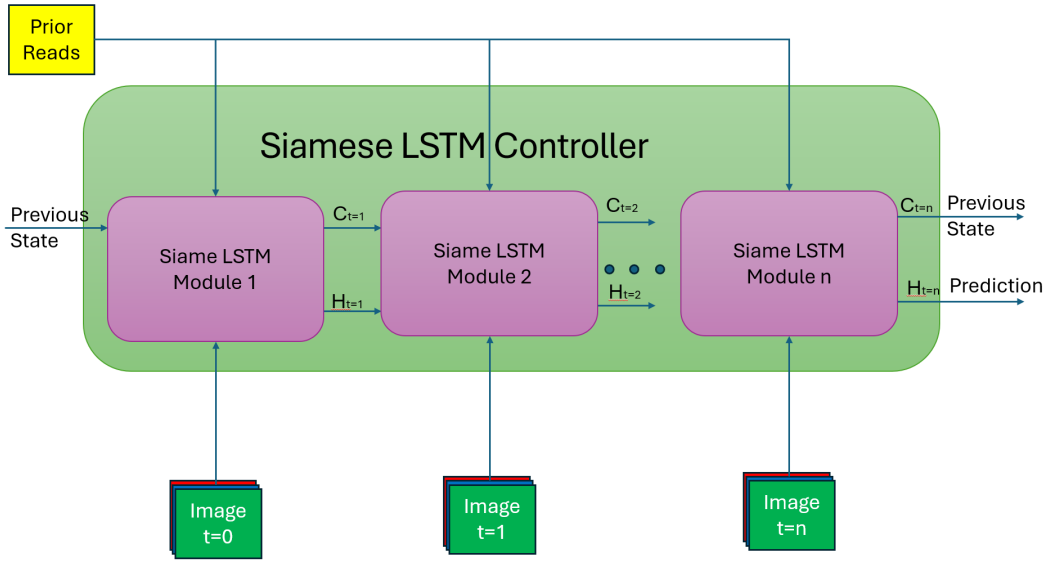}
  \caption{Depicts the many-to-one LSTM controller, which takes an image sequence, prior state, and prior reads, perform the input encoding.}
  \label{SLSTM_COntroller_FULL}
\end{subfigure}

\begin{subfigure}[t]{\textwidth}  
  \centering
  \includegraphics[width=\textwidth, trim={0 0 0 0.25cm}, clip]{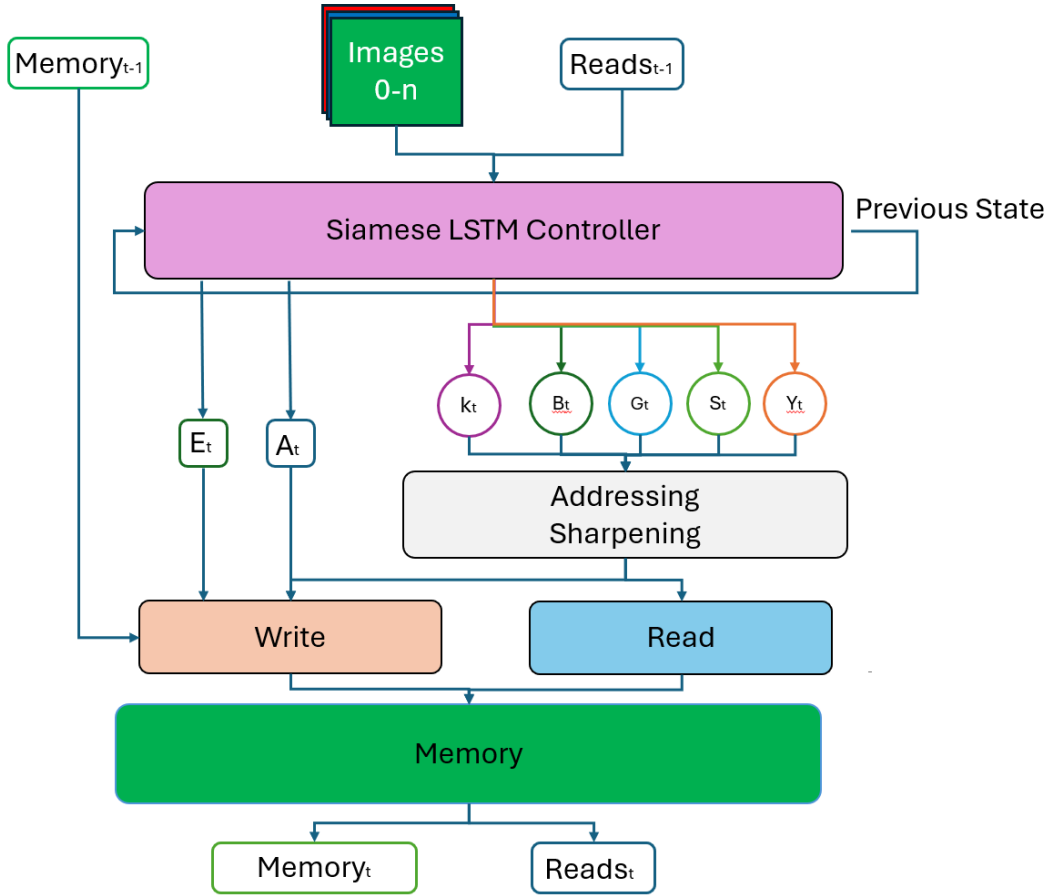} 
  \caption{Demonstrates the full Siamese Memory Augmented Network with the inclusion of the Siamese Controller allowing it to compare behavior.}
  \label{Siamese_MANN_FULL}
\end{subfigure}
\caption{An overview of the Siamese LSTM Controller and MANN.}
\vspace{-0.8cm}
\end{figure}

\begin{figure}[h]
  \centering
  \includegraphics[width=\textwidth, trim={0 0 0 0.5cm}, clip]{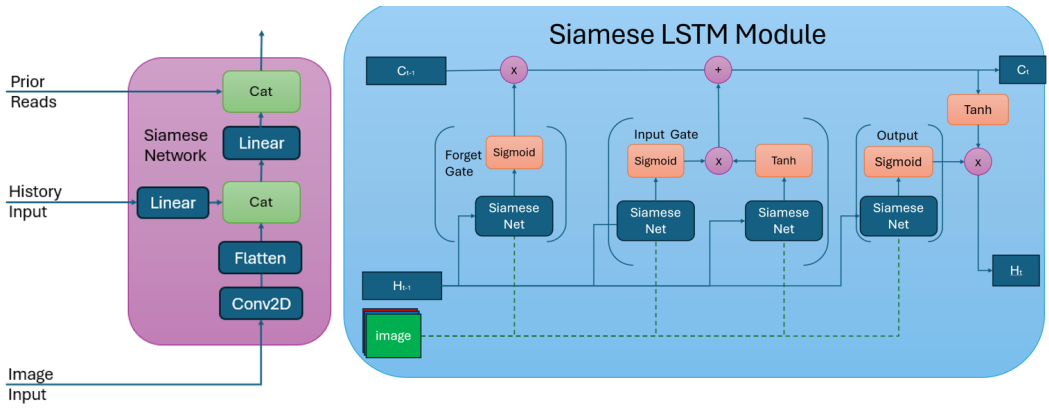}
  \caption{Demonstrates the Siamese Network that is then used as a gate for the SIAMESE LSTM Module, the inclusion of this network for the gates allows for the mixing of images and vectors}
  \label{SLSTM_CELL_FULL}
\end{figure}

\begin{figure}[!h]
    \centering
    \includegraphics[width=\textwidth, trim={0 0 0 0.7cm}, clip]{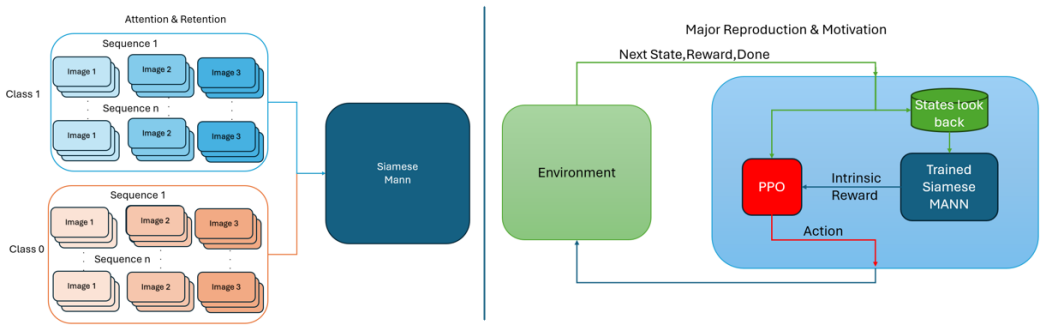} 
    \caption{Depicts the social learning frameworks where attention and retention use low-shot learning to train the Siamese MANN while reproduction and motivation occur through intrinsic rewards.}
    \label{Full_Social_FULL}
\end{figure}

\newpage
\subsection{Algorithmic Formulations}

\begin{algorithm}
\caption{Major Reproduction and Reinforcement}
\begin{algorithmic}[1]
    \Procedure{Training PPO with SMANN }{$Input$}=Trained SMANN
    \State Initialize Environment
    \State Initialize PPO
    \State $\beta$-Value
    \State Initialize Memory Buffer
    \State Initialize Size-n transition SMANN Buffer
    \State For 1000 Episodes
    \State \hspace{1em} Reset Environment
    \State \hspace{1em}Done set to False
    \State \hspace{1em}While Not Done:
    \State \hspace{1em} \hspace{1em} PPO Predicts Action
    \State \hspace{1em} \hspace{1em} Environment Returns Next-State,action,reward,done
    \State\hspace{1em} \hspace{1em} Append Next-State to SMANN Buffer
    \State\hspace{1em} \hspace{1em} SMANN Foward(SMANN Buffer)
    \State\hspace{1em} \hspace{1em}\hspace{1em} returns : bad-behavior-prob
    \State \hspace{1em} \hspace{1em}Intrinsic reward= bad-behavior-prob*$\beta$-Value
    \State \hspace{1em} \hspace{1em}Agent Reward=(Intrinsic reward+Extrinsic reward)
    \State \hspace{1em} \hspace{1em}Buffer append (state,next state,Agent Reward,done)
    \State\hspace{1em} \textbf{if} $steps == PPO update rate$ \textbf{then}
    \State \hspace{1em} \hspace{1em} Update PPO on Memory Buffer
    \EndProcedure
\end{algorithmic}
\label{Psuedo_Code_2 }
\caption{The pseudo code represents the training PPO with trained SMANN, which corresponds with the Major Reproduction and Reinforcement portion of social learning.}
\end{algorithm}

\begin{algorithm}
\caption{Attention and Retention}
\begin{algorithmic}[1]
    \Procedure{Training SMANN}{$Input$}
        \State \textbf{Given} $\beta_{parent}$ Dataset  
        \State set loss CrossEntropyLoss
        \State Initialize SMANN:
            \State \hspace{1em} Initialize Siamese Controller(S-LSTM)
            \State \hspace{1em} Initialize Read and Write Heads
            \State \hspace{1em} Initialize Memory 
            \State \hspace{1em} $Prediction_Layer$(fully connected) 
        \State Pad Previous state with zeros:
        \State \hspace{1em} Zero Pad: Previous Reads, Previous Controller State, Head State
        \State For  300 epochs 
        \State \hspace{1em} For Batches State Transition in $\beta_{parent}$ Dataset  
        \State \hspace{1em}\hspace{1em} SMANN Forward:
        \State \hspace{1em}\hspace{1em} \hspace{1em} S-LSTM Forward:
        \State \hspace{1em}\hspace{1em} \hspace{1em}\hspace{1em} Input:$\beta$,$P-reads$,$P-ControllerState$
        \State \hspace{1em}\hspace{1em} \hspace{1em}\hspace{1em} returns:Encoding,C-state
        \State \hspace{1em}\hspace{1em} \hspace{1em}Using Read-Heads(Encoding,Write,usage,least usage) 
        \State \hspace{1em}\hspace{1em} \hspace{1em}\hspace{1em} returns :Reads-from-Memory
        \State \hspace{1em}\hspace{1em} \hspace{1em}Using Write-Heads(Encoding,Write,usage,least usage)  
        \State \hspace{1em}\hspace{1em} \hspace{1em}\hspace{1em} returns :Written-Memory
        \State \hspace{1em}\hspace{1em} \hspace{1em} Logits=Cat(Encoding,reads-from-memory)
        \State \hspace{1em}\hspace{1em} \hspace{1em} Softmax(Prediction-Layer(Out)) 
        \State \hspace{1em}\hspace{1em} \hspace{1em}\hspace{1em} returns class-prediction 
        \State \hspace{1em}\hspace{1em} CrossEntropyLoss($\beta$-predicted,$\beta$-true) returns loss
        \State \hspace{1em}\hspace{1em} \hspace{1em} returns loss
        \State \hspace{1em}\hspace{1em} optimization step
    \EndProcedure
\end{algorithmic}
\label{Psuedo_Code_2 }
\caption{The pseudo code represents the training of the SMANN algorithm, which corresponds with the attention and retention portion of social learning.}
\end{algorithm}

\newpage

\subsection{Hyperparameters}

\begin{table}[h!]
  \caption{Method Parameters}
  \label{Method_Params}
  \centering
  \begin{tabular}{lll}
    \toprule
    \multicolumn{2}{c}{PPO}                   \\
    \cmidrule(r){1-2}
    Name          & Amount \\
    \midrule
    learning rate     &$1e-5$    \\
    update rate       & $350 steps$\\
    epochs              &40  \\
    \bottomrule \\
    \multicolumn{2}{c}{MANN and SMANN}                   \\
    \cmidrule(r){1-2}
    Name          & Amount \\
    learning rate   &$1e-3$    \\
    Epochs MANN     &300\\
    Epochs SMANN    &150\\
    Siamese Controller Layer &3 \\
    Controller Layer    &3\\
    Read Head           &10\\
    Read write          &10\\
    Memory M      &$128$\\
    Memory N         &$40$\\
    \bottomrule
  \end{tabular}
  \caption{This table shows the hyperparameters for PPO, MANN, and SMANN. All hyperparameters that vary from SMANN to MANN are independently labeled.}
\end{table}

\subsection{PPO Baseline Graphs}

\vspace{-7.0cm}

\begin{figure}[h]
    \centering
    \medskip
    \begin{subfigure}[b]{0.9\textwidth}
        \centering
        \includegraphics[width=\textwidth, trim={0 0 0 1.2cm}, clip]{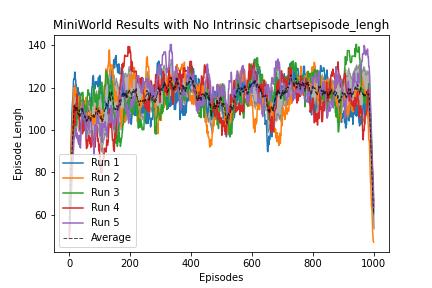}
        \caption{Episode Length}
        \label{fig:BASE_LENGHT}
    \end{subfigure}
    \hfill
    \begin{subfigure}[b]{0.9\textwidth}
        \centering
        \includegraphics[width=\textwidth, trim={0 0 0 1.2cm}, clip]{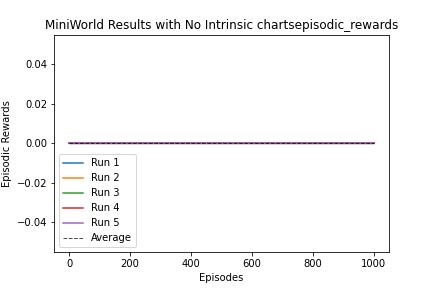}
        \caption{Episodic Reward}
        \label{fig:BASE_EXTRINSIC}
    \end{subfigure}
    \caption{These Results demonstrate the achieved episodic reward of base PPO and the achieved episode length}
    \label{fig:BASE_RESULTS}
\end{figure}

\subsection{Larger Results Figures}

\begin{figure}[H]
   \centering
    \begin{subfigure}[t]{\textwidth}
        \centering
        \includegraphics[width=0.7\textwidth, trim={0 0 0 1.2cm}, clip]{Images/Results/Baseline_Intrinsic_Stimuli/Episode_Lengh/MiniWorld_Results_with_Stimuli_Intrinsic_chartsepisode_lengh.jpeg}
        \caption{Episode Length}
        \label{fig:STIMULI_LENGHT}
    \end{subfigure}
    \begin{subfigure}[t]{\textwidth}
        \centering
        \includegraphics[width=0.7\textwidth, trim={0 0 0 1.1cm}, clip]{Images/Results/Baseline_Intrinsic_Stimuli/Extrinsic_Reward/MiniWorld_Results_with_Stimuli_Intrinsic_chartsepisodic_rewards.jpeg}
        \caption{Episodic Reward}
        \label{fig:STIMULI_EXTRINSIC}
    \end{subfigure}
    \begin{subfigure}[t]{\textwidth}
        \centering
        \includegraphics[width=0.7\textwidth, trim={0 0 0 1.2cm}, clip]{Images/Results/Baseline_Intrinsic_Stimuli/Intrinsic_Reward/MiniWorld_Results_with_Stimuli_Intrinsic_chartsintrinsic_rewards.jpeg}
        \caption{Episodic Intrinsic}
        \label{fig:STIMULI_INTRINSIC}
    \end{subfigure}
    \caption{The stimuli fear's constant negative reward approximates a living cost, making it optimal for the agent to go to the terminal condition, promoting a decrease in episode length. }
    \label{fig:STIMULI_RESULTS_FULL}
\end{figure}

\begin{figure}[H]
    \centering
    \medskip
    \begin{subfigure}[t]{\textwidth}
        \centering
        \includegraphics[width=0.8\textwidth, trim={0 0 0 1.2cm}, clip]{Images/Results/Thresholded_Intrinsic_Behavior/Threshold_Numbersteps_Per_Episode/MiniWorld_Results_with_Expanded_Controllerand_thresholded0.25_chartsepisode_lengh.jpeg}
        \caption{Episode Length}
        \label{fig:LOW_LENGHT}
    \end{subfigure}
    \begin{subfigure}[t]{\textwidth}
        \centering
        \includegraphics[width=0.8\textwidth, trim={0 0 0 1.2cm}, clip]{Images/Results/Thresholded_Intrinsic_Behavior/Threshold_Extrinsic_Reward/MiniWorld_Results_with_Expanded_Controller_and_thresholded0.25_chartsepisodic_rewards.jpeg}
        \caption{Episodic Reward}
        \label{fig:LOW_EXTRINSIC}
    \end{subfigure}
    \begin{subfigure}[t]{\textwidth}
        \centering
        \includegraphics[width=0.8\textwidth, trim={0 0 0 1.2cm}, clip]{Images/Results/Thresholded_Intrinsic_Behavior/Threshold_Intrinsic_Reward/MiniWorld_Results_with_Expanded_Controller_and_thresholded0.25_chartsintrinsic_rewards.jpeg}
        \caption{Episodic Intrinsic}
        \label{fig:LOW_INTRINSIC}
    \end{subfigure}
    \caption{At a low threshold value, the intrinsic reward punishes any policy that approximates the behavior promoting the longest episode length.}
    \label{fig:LOW_RESULTS_FULL}
\end{figure}

\begin{figure}[H]
    \centering
    \medskip
    \begin{subfigure}[b]{\textwidth}
        \centering
        \includegraphics[width=0.8\textwidth, trim={0 0 0 1.2cm}, clip]{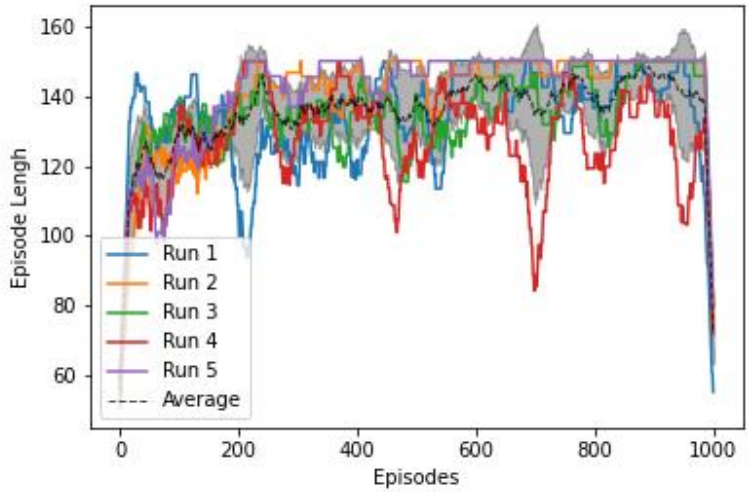}
        \caption{Episode Length}
        \label{fig:MID_LENGHT}
    \end{subfigure}
    \begin{subfigure}[b]{\textwidth}
        \centering
        \includegraphics[width=0.8\textwidth, trim={0 0 0 1.2cm}, clip]{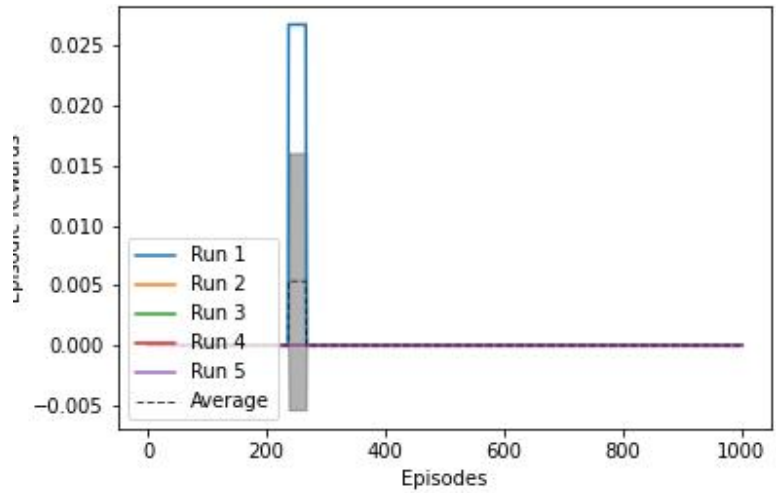}
        \caption{Episodic Reward}
        \label{fig:MID_EXTRINSIC}
    \end{subfigure}
    \begin{subfigure}[b]{\textwidth}
        \centering
        \includegraphics[width=0.8\textwidth, trim={0 0 0 1.2cm}, clip]{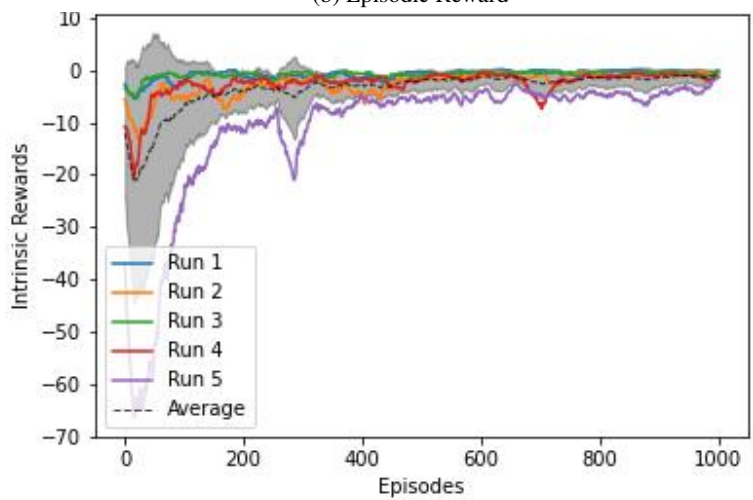}
        \caption{Episodic Intrinsic}
        \label{fig:MID_INTRINSIC}
    \end{subfigure}
    \caption{Mid threshold fear promotes a policy that neither optimizes for greater episode length through safe exploration or greater exploration through riskier actions.}
    \label{fig:BEHAVIOR_MID_RESULTS_FULL}
\end{figure}

\begin{figure}[H]
    \medskip
    \begin{subfigure}[b]{\textwidth}
        \centering
        \includegraphics[width=0.8\textwidth, trim={0 0 0 1.2cm}, clip]{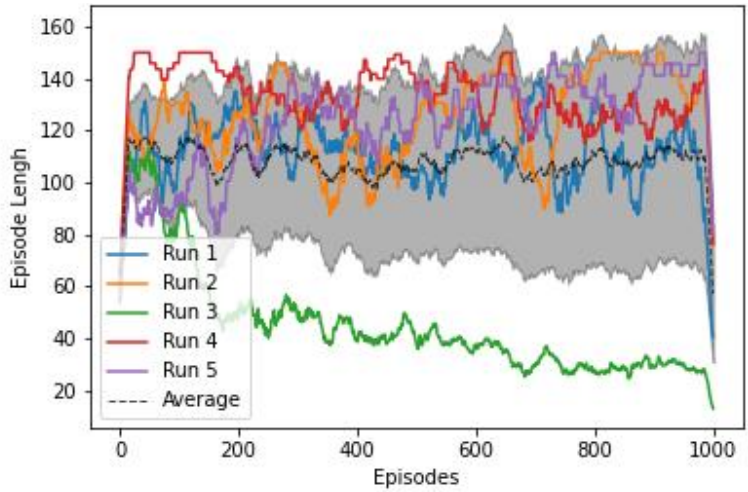}
        \caption{Episode Length}
        \label{fig:HIGH_LENGHT}
    \end{subfigure}
    \begin{subfigure}[b]{\textwidth}
        \centering
        \includegraphics[width=0.8\textwidth, trim={0 0 0 1.2cm}, clip]{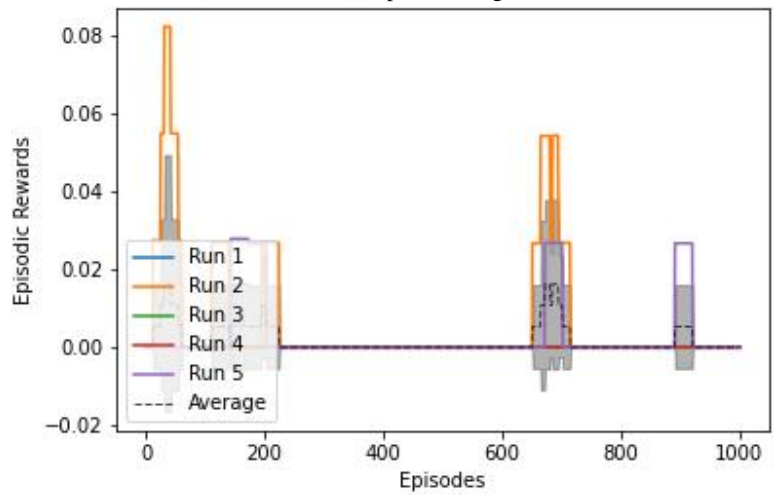}
        \caption{Episodic Reward}
        \label{fig:HIGH_EXTRINSIC}
    \end{subfigure}
    \begin{subfigure}[b]{\textwidth}
        \centering
        \includegraphics[width=0.8\textwidth, trim={0 0 0 1.2cm}, clip]{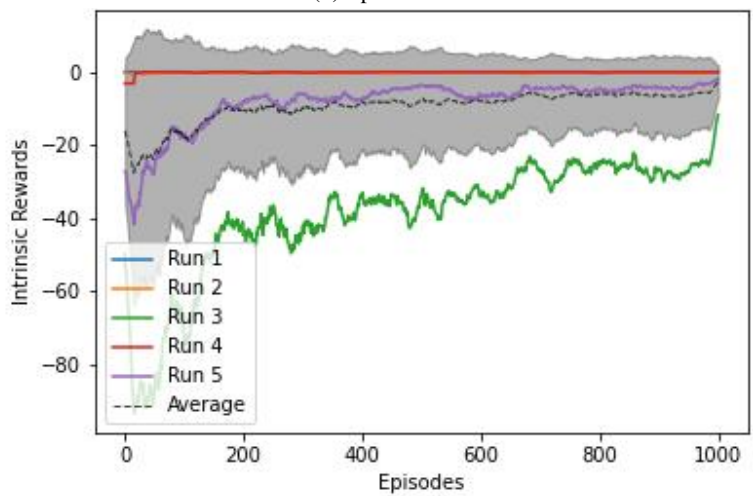}
        \caption{Episodic Intrinsic}
        \label{fig:HIGH_INTRINSIC}
    \end{subfigure}
    \caption{The High fear threshold allows for the most exploration, increasing while still punishing the pristine representation of fear, increasing the probability of success, but reducing episode length}
    \label{fig:BEHAVIOR_HIGH_RESULTS_FULL}
\end{figure}

\subsection{Qualitative Results}
\begin{figure}[H]
   \centering
    \begin{subfigure}[t]{0.32\textwidth}
        \centering
        \includegraphics[width=0.9\textwidth, trim={0 0 0 1.2cm}, clip]{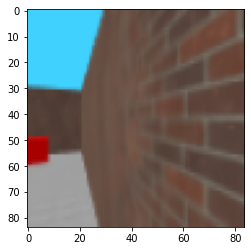}
        \caption{Step 99}
    \end{subfigure}
   \centering
    \begin{subfigure}[t]{0.32\textwidth}
        \centering
        \includegraphics[width=0.9\textwidth, trim={0 0 0 1.2cm}, clip]{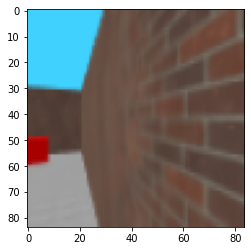}
        \caption{Step 100}
    \end{subfigure}
       \centering
    \begin{subfigure}[t]{0.32\textwidth}
        \centering
        \includegraphics[width=0.9\textwidth, trim={0 0 0 1.2cm}, clip]{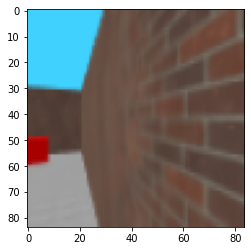}
        \caption{Step 101}
    \end{subfigure}
       \centering
    \begin{subfigure}[t]{0.32\textwidth}
        \centering
        \includegraphics[width=0.9\textwidth, trim={0 0 0 1.2cm}, clip]{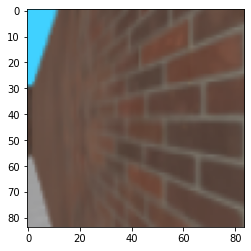}
        \caption{Step 102}
    \end{subfigure}
       \centering
    \begin{subfigure}[t]{0.32\textwidth}
        \centering
        \includegraphics[width=0.9\textwidth, trim={0 0 0 1.2cm}, clip]{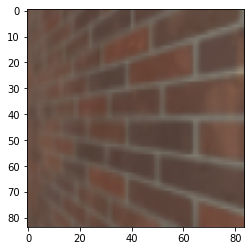}
        \caption{Step 103}
    \end{subfigure}
    \caption{This is an example of the .25 threshold fifth run where the agent moves towards a zero intrinsic reward state, avoiding any representation of danger }
    \label{fig:GAD_VISUAL}
\end{figure}

\end{document}